\documentclass[times,twocolumn,final]{elsarticle}

\usepackage{ipcai_arxiv}
\usepackage{framed,multirow}

\usepackage{amssymb}
\usepackage{latexsym}

\usepackage{url}
\usepackage{cite}
\usepackage{amsmath,amssymb,amsfonts,bm}
\usepackage{booktabs}
\usepackage{arydshln}
\usepackage{tabulary}
\usepackage{graphicx,epstopdf}
\usepackage{textcomp}
\usepackage{subcaption}
\usepackage{booktabs}
\usepackage[svgnames,table]{xcolor}
\usepackage{pifont}
\usepackage{arydshln}
\usepackage[colorlinks=true,citecolor=cyan,urlcolor=cyan]{hyperref}
\usepackage{graphicx}
\usepackage{placeins}
\usepackage{fancyhdr}
\usepackage{float}

\fancypagestyle{firstpagestyle}{
    \fancyhf{} 
    \fancyhead{} 
    \fancyfoot{} 
    \fancyhead[CO]{\em \fontsize{9pt}{8pt}\selectfont This article has been accepted to the International Conference on Information Processing in Computer-Assisted Interventions, 2025.}
}

\begin{document}


\title{When do they StOP?: A First Step Towards Automatically Identifying Team Communication in the Operating Room}

\author[1]{Keqi \snm{Chen}\corref{cor}}
\author[3,7]{Lilien \snm{Schewski}}

\author[1,2]{Vinkle \snm{Srivastav}}

\author[2,4,5]{Joël \snm{Lavanchy}}

\author[2,6]{Didier \snm{Mutter}}

\author[3,7]{Guido \snm{Beldi}}

\author[3,7]{Sandra \snm{Keller}\corref{co-last}}

\author[1,2]{Nicolas \snm{Padoy}\corref{co-last}}

\cortext[cor]{Corresponding author: keqi.chen@unistra.fr}
\cortext[co-last]{shared last authorship}

\address[1]{University of Strasbourg, CNRS, INSERM, ICube, UMR7357, France}

\address[2]{IHU Strasbourg, Strasbourg 67000, France}

\address[3]{Department for Biomedical Research (DBMR), University of Bern, 3008 Bern, Switzerland}

\address[4]{University Digestive Health Care Center, Clarunis, Basel 4002, Switzerland}

\address[5]{Department of Biomedical Engineering, University of Basel, Allschwil 4123, Switzerland}

\address[6]{University Hospital of Strasbourg, Strasbourg 67000, France}

\address[7]{Department for Visceral Surgery and Medicine, Bern University Hospital, University of Bern, 3010 Bern, Switzerland}

\received{XXX}
\finalform{XXX}
\accepted{XXX}
\availableonline{XXX}
\communicated{XXX}

\begin{abstract}

\textbf{Purpose: } Surgical performance depends not only on surgeons' technical skills but also on team communication within and across the different professional groups present during the operation. Therefore, automatically identifying team communication in the OR is crucial for patient safety and advances in the development of computer-assisted surgical workflow analysis and intra-operative support systems. To take the first step, we propose a new task of detecting communication briefings involving all OR team members, i.e. the team Time-out and the StOP?-protocol, by localizing their start and end times in video recordings of surgical operations.

\textbf{Methods: } We generate an OR dataset of real surgeries, called Team-OR, with more than one hundred hours of surgical videos
captured by the multi-view camera system in the OR. The dataset contains temporal annotations of 33 Time-out and 22 StOP?-protocol activities in total. We then propose a novel group activity detection approach, where we encode both scene context and action features, and use an efficient neural network model to output the results.

\textbf{Results: } The experimental results on the Team-OR dataset show that our approach outperforms existing state-of-the-art temporal action detection approaches. It also demonstrates the lack of research on group activities in the OR, proving the significance of our dataset. 

\textbf{Conclusion: } We investigate the Team Time-Out and the StOP?-protocol in the OR, by presenting the first OR dataset with temporal annotations of group activities protocols, and introducing a novel group activity detection approach that outperforms existing approaches. Code is available at \url{https://github.com/CAMMA-public/Team-OR}.
\end{abstract}

\maketitle
\thispagestyle{firstpagestyle}

\section{Introduction}\label{intro}

The operating room (OR) is a fast-paced high-stakes socio-technical environment involving communication between different professional groups (i.e. OR teams), including surgeons, nurses, and anesthesiologists.  Communication in OR teams is essential for good team performance as it largely depends on effective team communication and interactions during surgical procedures~\citep{mazzocco2009surgical}. In order to build modern context-aware OR support systems of computer-assisted interventions, automatic analysis and recognition of activities performed by OR teams is crucial~\citep{padoy2019machine,mascagni2021or}. 

\begin{figure*}[t]
\centering
\includegraphics[width=0.9\textwidth]{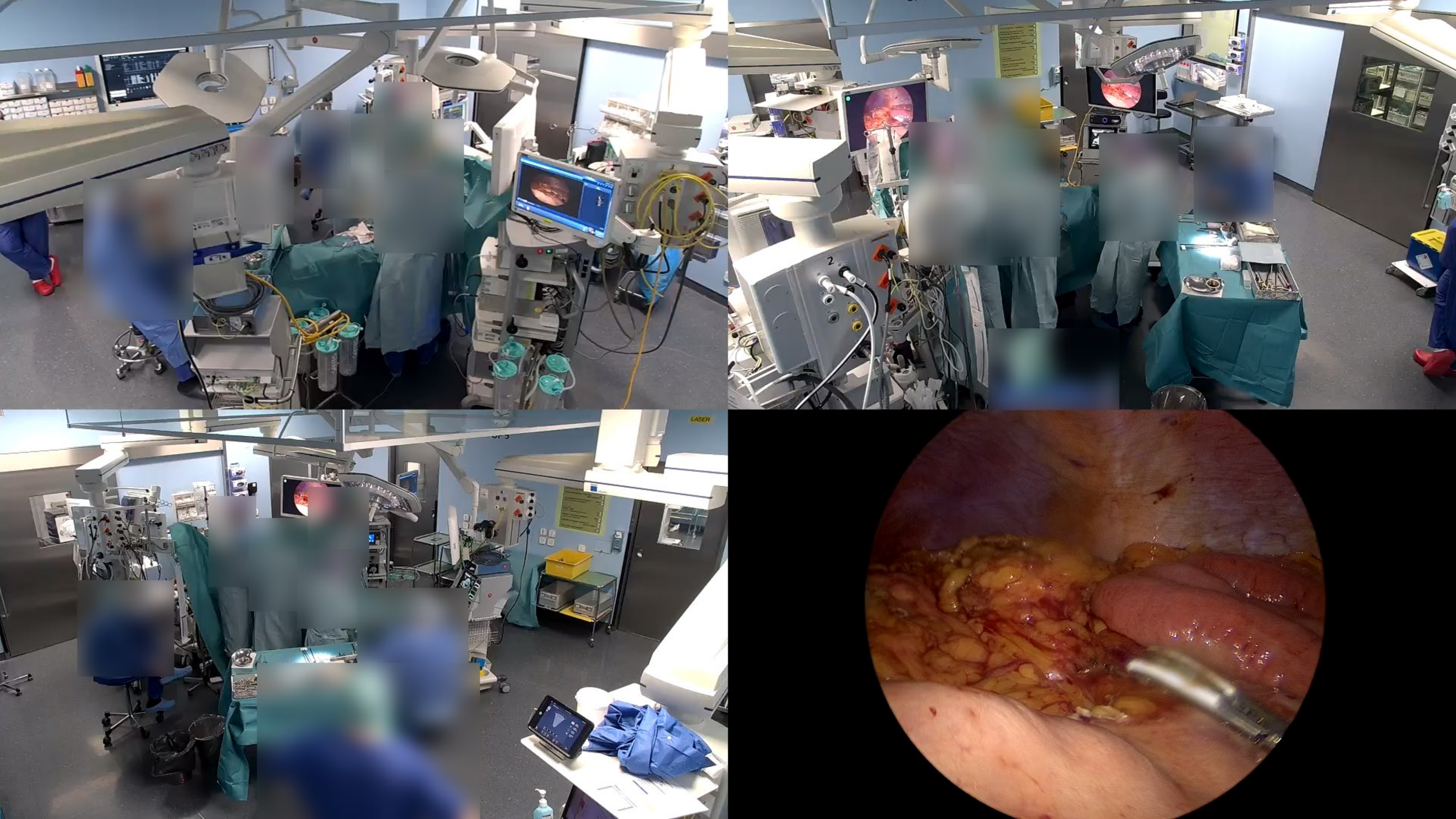}
\caption{Overview of the Team-OR dataset, consisting of synchronized three ceiling camera views and one laparoscopic view. We blurred the half bodies of the team for privacy concern.}\label{fig:dataset}
\end{figure*}

Research showed that team communication briefings in healthcare teams, including in OR teams, are important contributors to patient safety. We focus on two such briefings, the team Time-out and the StOP?-protocol. (1) A Time-out usually happens before the first incision, at the very beginning of the operation. During the Time-Out, OR team members review together essential information about the surgical procedure, including the patient's identity, operative site, surgical procedure, patient allergies, and voice potential concerns~\citep{johnston2009surgical}. The Time-Out is a standard part of most surgeries in the world and is supported by the WHO \citep{haynes2009surgical}. (2) The StOP?-protocol is initiated by the main surgeon during the surgical procedure~\citep{tschan2022effects}. All OR team members pause their taskwork and focus on team communication for about 30-90 seconds. All team members present are informed about the status of the operation (St), the objectives (O) of the next steps of the operation, potential problems (P), and questions (?) that can be voiced. The StOP?-protocol was implemented in several hospitals as part of a first study~\citep{tschan2022effects} and is currently being tested as part of a large randomized controlled trial~\citep{keller2022stop}. There is evidence that both team Time-outs and the StOP?-protocol contribute to improving patient safety (e.g. diminish wrong site surgery)~\citep{haynes2009surgical} and patient clinical post-operative outcomes such as mortality, unplanned reoperations, and length of hospital stay~\citep{tschan2022effects,van2012effects}. The implementation of such briefings is a challenge in many healthcare systems, as it disrupts the established team routines, requiring training and continuous process monitoring~\citep{beldi2009impact,kurmann2014impact,seelandt2014assessing}. It has been shown that video-based technologies can support these implementations~\citep{overdyk2016remote}. Therefore, investigating such group activities is a promising first step toward automatic team interaction analysis in the OR. 

In recent years, a few OR datasets have been proposed~\citep{belagiannis2016parsing,srivastav2018mvor,ozsoy20224d}, which concentrate on human pose estimation and scene graph generation. Although these datasets provide solid foundations for atomic-level human behavior analysis, the long-range group activity analysis in the OR remains unexplored. Moreover, the MVOR dataset~\citep{srivastav2018mvor} only contains selected images, and the 4D-OR dataset~\citep{ozsoy20224d} is composed of simulated surgeries, which are much shorter and do not contain the complex team interactions of real surgeries. Therefore, relevant OR datasets containing actual surgical procedures with clinically validated team interaction protocols along with the strong baseline methods will be instrumental for the holistic understanding of the OR.


To this end, we generate the multi-view OR dataset of real abdominal surgeries with team interaction annotations, called Team-OR (Operating Room dataset for Team activity analysis), as shown in Fig.~\ref{fig:dataset}. The dataset is recorded at a University Hospital in Western Europe and consists of videos of 37 laparoscopic surgeries, with a total duration of 105 hours, and temporal annotations of 33 ``Time-out'' and 22 ``StOP?''. The videos were recorded as a subsample of operations included in the StOP? II trial \citep{keller2022stop}.   
We propose a novel group activity detection approach to automatically detect the start- and end-time of Time-out and StOP? activities in the untrimmed videos. We encode both global scene visual features and local skeleton-based features through VideoMAEv2~\citep{wang2023videomae} and STGCN++~\citep{duan2022pyskl} models. Afterward, we propose an efficient neural network model for group activity detection, where we construct effective multi-level features through Max and Average Poolings. Experiments on the Team-OR dataset prove the effectiveness of our approach. 

Our contributions can be summarized as follows: (1) We highlight the importance of automatic team interaction analysis, and generate a multi-view OR dataset of real surgeries with team interaction annotations, Team-OR; (2) We present a novel group activity detection approach to detect the ``Time-out'' and ``StOP?'' activities, which obtain state-of-the-art performance. 

\section{Related work}\label{related}

\noindent \textbf{Operating room dataset: } Belagiannis \textit{et al.}~\citep{belagiannis2016parsing} propose the first multi-view OR dataset of simulated surgeries with 3d human pose annotations. To introduce data captured during real interventions, Srivastav \textit{et al.} propose the MVOR image dataset, the first multi-view RGB-D dataset with 3d human poses. Then, to untangle the interactions between clinicians and objects, Özsoy \textit{et al.}~\citep{ozsoy20224d} propose the 4D-OR dataset of simulated knee surgeries with fine-grained semantic scene graph annotations. With the existing OR datasets, several computer vision tasks in the OR are supported such as human pose estimation~\citep{hansen2019fusing,srivastav2022unsupervised}, semantic scene graph generation~\citep{ozsoy2023labrad,pei2024s} and surgical phase recognition~\citep{ozsoy2024holistic}. However, none of these datasets record real team interactions during surgeries, and thus do not support the analysis of the team Time-out and the StOP?-protocol.


\noindent \textbf{Temporal action detection: } Temporal action detection (TAD) is a classical computer vision task, which aims to localize and classify all the actions in an untrimmed video. Compared to temporal segmentation (TS), TAD is more challenging due to issues such as data imbalance and more ambiguous action boundaries. The early TAD approaches usually follow a two-stage design, where they conduct proposal generation and classification separately~\citep{lin2019bmn,xu2020g}. In order to simplify the pipeline for end-to-end training, one-stage approaches also become popular by simultaneously localizing and classifying the actions~\citep{lin2021learning,zhang2022actionformer}. In the medical area, although there are works studying the TAD in the cataract and nephrectomy surgical videos~\citep{hao2023act,chandra2024vitals,luo2024surgplan} and TS in both endoscopic and OR videos~\citep{quellec2014real,dergachyova2016automatic,volkov2017machine,ozsoy20224d}, the field of TAD in the OR with clinicians as the main actors remains unexplored.

\section{Methodology}\label{method}

\subsection{Team-OR dataset}

\begin{figure*}[h]
\centering
\includegraphics[width=0.9\textwidth]{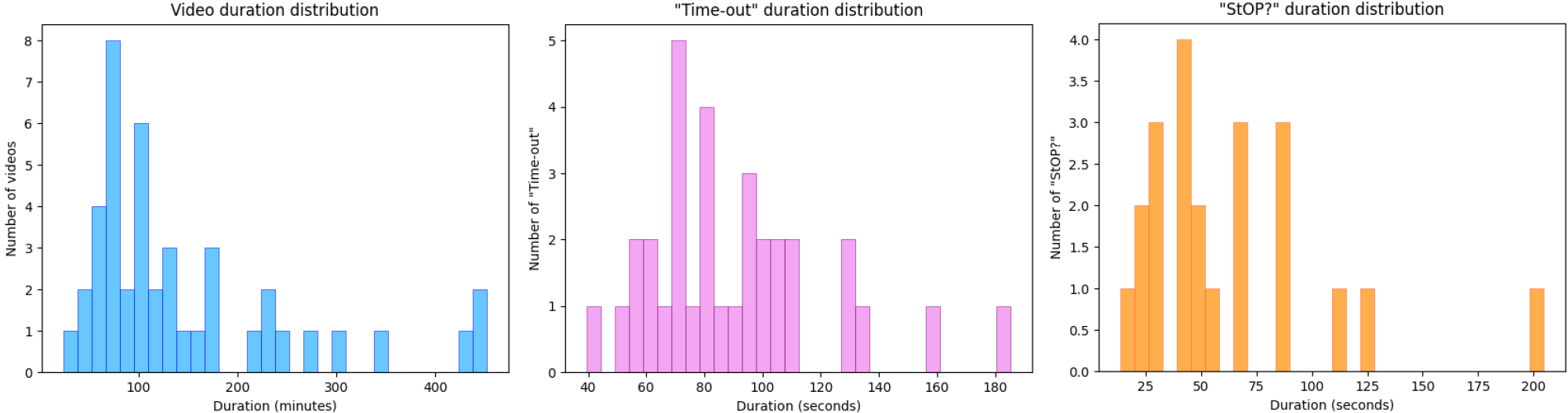}
\caption{The duration distribution of the videos, ``Time-out'' and ``StOP?''.}\label{fig:statistics}
\end{figure*}

In order to study the team communications during real surgeries in the OR, we introduce the Team-OR dataset. The Team-OR dataset consists of untrimmed operation videos of 37 laparoscopic surgeries, captured by three ceiling cameras and one laparoscopic camera inside the OR. These cameras are stationary and synchronized, which can cover most of the OR area. The videos consisted of bariatric, bile duct, colon, liver, gastric, small bowel, thyroid, rectal, esophagus, gallbladder, and herniorrhaphy surgical operations, covering a wide variety of abdominal surgical procedures. In most cases, the videos start after patients are ready and before the Time-out and end after the surgery is finished. Thus, one operation corresponds to one untrimmed video. However, some recordings of the operations needed to be paused for a moment when someone who had not given consent entered the OR. Therefore, we have 43 videos at 30 frames-per-second (FPS) altogether, with a total duration of over 105 hours. The distribution of the video duration is shown in Fig.~\ref{fig:statistics}. 

Since the Time-out and StOP?-protocol are communication briefings performed during operations included in the video, we annotate the start- and the end-time of such group activities in every video through the MOSaiC platform~\citep{mazellier2023mosaic}. We manually annotate these activities using both vision and audio signals, with boundary errors of less than five seconds. It is to be noted that we cannot use the audio after annotating for privacy concerns. In the end, we have 33 Time-outs and 22 StOP?-protocol annotations, with an average duration of 89.8 and 62.9 seconds respectively. We show the distribution of these activities' duration in Fig.~\ref{fig:statistics}. 

\subsection{Hierarchical analysis of group activities}\label{hie}

Given untrimmed OR videos during surgeries, our goal is to localize all the Time-out and StOP?-protocol activities, namely group activity detection. Although the task shares a similar definition of general TAD, it differs in two aspects: (1) we only have a few positive examples of the activities among a large number of negative samples; (2) the visual differences between such group activities and normal situation are very marginal, especially the StOP?-protocol. Therefore, it is significant to analyze the key features of the group activities hierarchically. 

\begin{figure*}[h]
\centering
\includegraphics[width=0.9\textwidth]{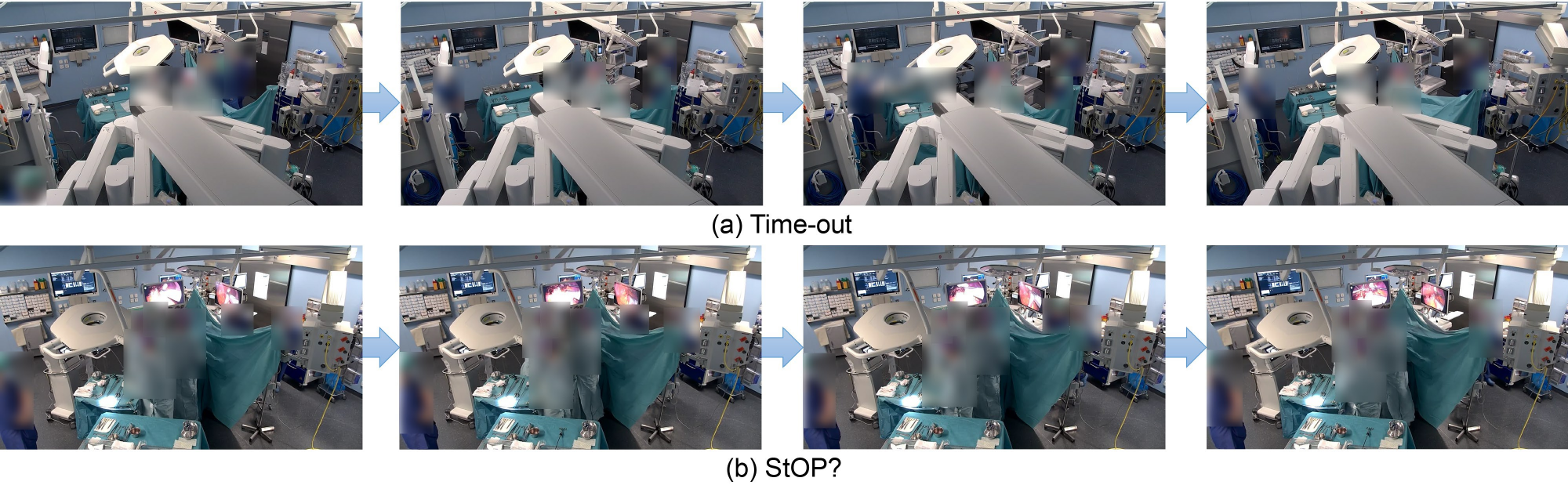}
\caption{Examples of the Time-out and StOP?-protocol activities in the dataset. We blurred the half-bodies of the team for privacy concerns.}\label{fig:example}
\end{figure*}

As shown in Fig.~\ref{fig:example}, for a proper implementation of both protocols, every member of the OR team should actively participate by paying attention and stopping their work for some seconds; for the StOP?-protocol, OR team members are instructed to approach the OR table and stop any manual activity \citep{papadakis2019safer,tschan2022effects}.  However, the problem of poor compliance leads to some briefings not being carried out correctly, typically when OR team members do not completely stop manual activities. This makes the task difficult, as team members' positions are not always reliable features for localizing these activities. 
In order to have an in-depth analysis by comparing the positive samples with hard negative samples, we split the OR video into clips of 30-second duration and train a spatial-temporal graph convolution network model to classify these clips by encoding the clinicians' spatial-temporal locations. Correspondingly, we use False Positive Rate at 95\% recall (FPR-95) to evaluate the performance. Although the model performs poorly with over 90\% FPR-95 for StOP? and 40\% FPR-95 for Time-out, we still sort the false positive samples by confidence, and manually compare the hardest cases with the real activities to find the effective visual features. Finally, we draw hierarchical observations of the team interactions as follows: 

\begin{enumerate}
    \item During the Time-out,  some team members are usually preparing the necessary devices and instruments, with their poses continuously changing. Therefore, the unprepared scene context and the persons' skeleton sequences are important. 
    \item During the StOP?, the team members gather around the operating table, and the main surgeon talks to them with a potential co-speech gesture. The others may nod, read documents, and share their opinions, but most of them slightly move and concentrate on the speaker. In the end, resumes taskwork. Therefore, the team members' non-verbal interactions are important. 
    \item Both activities' boundaries are visually vague.  
\end{enumerate}

To summarize, we need to encode both global scene context and local individual action features so that we can effectively model the patterns of the group activities. 

\subsection{Group activity detection}

\begin{figure*}[h]
\centering
\includegraphics[width=0.9\textwidth]{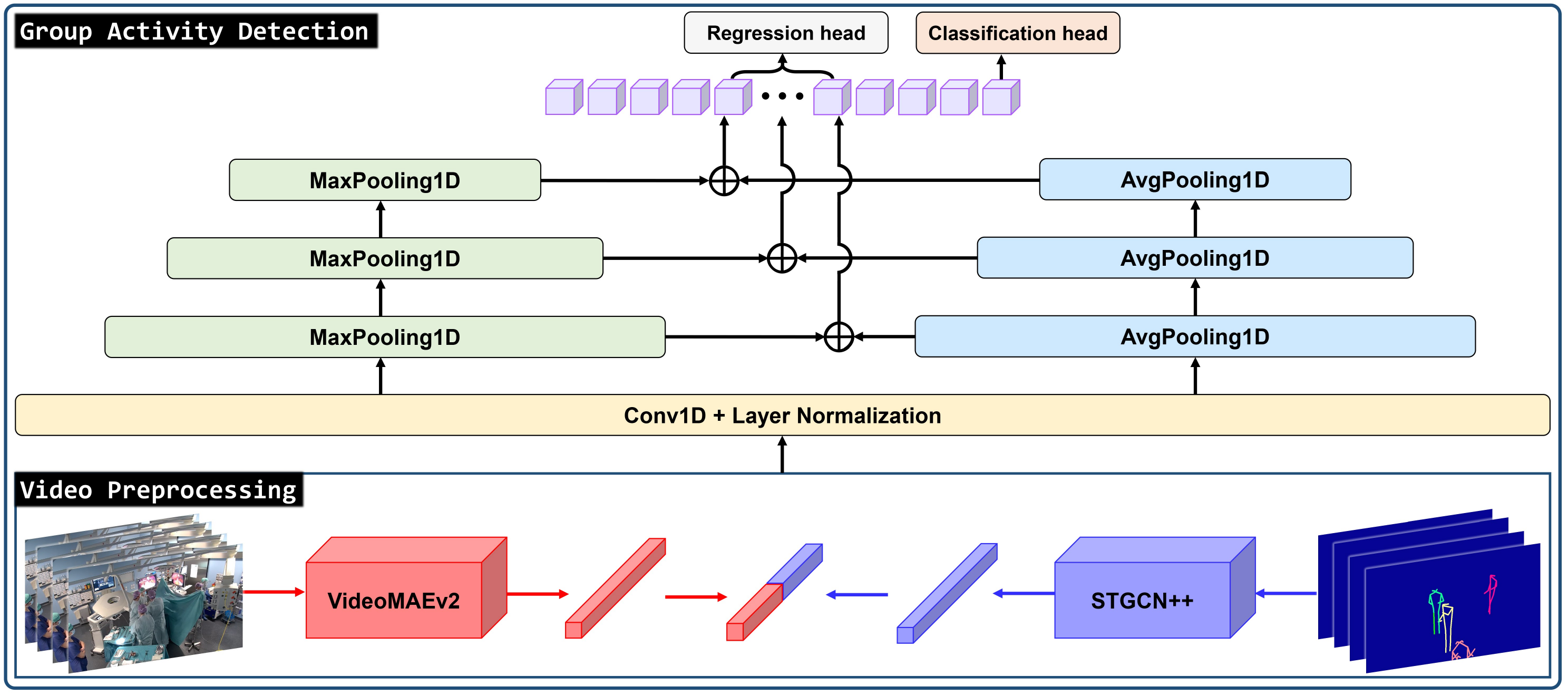}
\caption{Framework of our approach. We extract temporal scene context and skeleton features through pretrained VideoMAEv2 \citep{wang2023videomae} and STGCN++ \citep{duan2022pyskl} models, and then use a light-weight neural network model to detect the group activities in the OR. }\label{fig:method}
\end{figure*}

In this section, we introduce the pipeline of the proposed algorithm as shown in Fig.~\ref{fig:method}, where we utilize one camera view as the input. 

Since it is difficult to directly process the untrimmed operation videos, and we have only a few positive labels for training, it is necessary to first extract robust features of the videos using pretrained models for better generalization ability. As discussed in Section~\ref{hie}, we need to extract both global scene context and local individual features. For the holistic scene context, we utilize the pretrained VideoMAEv2 model~\citep{wang2023videomae} to extract the temporal features $F_g$. For the individuals, we first use the YOLOX detector~\citep{yolox2021} and the ByteTrack tracker~\citep{zhang2022bytetrack} to obtain the tracklets of all the team members during surgery. Then, we apply HRNet~\citep{sun2019deep} to predict the poses at every timestamp, and thus obtain the skeleton tracklets. Afterward, we use a pretrained STGCN++ model~\citep{duan2022pyskl} to extract the clinicians' action features $F_l$. Finally, for each clip unit, we concatenate its global and local features as the extracted video features $F$: 

\begin{equation}
F = (F_{g}, F_{l})
 \label{eq:fea}
\end{equation}

After video preprocessing, we propose an efficient neural network model for group activity detection. Specifically, we first pass the obtained input features $\mathcal{F}=\left\{ F_1, F_2, ..., F_T \right\}$ of size $T$ through the backbone, consisting of two 1D convolutional layers and layer normalization \citep{ba2016layer}. With the obtained feature $Z^1$, we use two separate branches of Max Poolings and Average Poolings respectively with stride 2, to generate two feature pyramids $Z_m$ and $Z_a$, without introducing extra parameters. For pyramids of $L$ layers, we compute each layer's features as follows, where $2 \leq l \leq L$:
\begin{equation}
Z_m^l=\mathrm{MaxPooling}(Z_m^{l-1})
 \label{eq:max}
\end{equation}
\begin{equation}
Z_a^l=\mathrm{AvgPooling}(Z_a^{l-1})
 \label{eq:avg}
\end{equation}
The idea of using Max Pooling comes from \citep{tang2023temporalmaxer}, and we extend it by adding another Average Pooling branch, which is proven to be useful in Section~\ref{sec:abl}. Afterward, we merge the two feature pyramids by adding the features of the same layer $l$:
\begin{equation}
Z^l=Z_m^l + Z_a^l
 \label{eq:max_avg}
\end{equation}
Lastly, we use a regression head and a classification head to predict the positive activity proposals along with the boundaries. Specifically, for every moment instant across $L$ layers, the classification head predicts the probability of the activity, and the regression head predicts the starting and ending point of the potential activity. During inference, Soft-NMS~\citep{bodla2017soft} is applied to remove duplicated activity proposals. 

During training, we use focal loss \citep{ross2017focal} and temporal DIoU loss \citep{zheng2020diou} as the classification loss $L_{\mathrm{cls}}$ and the regression loss $L_{\mathrm{reg}}$ respectively. The overall loss $L$ is calculated as follows, where $\sigma_{\mathrm{IoU}}$ is the temporal IoU between the proposal and the ground-truth activity sequence, and ${N_{\mathrm{neg}}}$ and ${N_{\mathrm{pos}}}$ are the number of negative and positive samples: 

\begin{equation}
L = \frac{1}{N_{\mathrm{pos}}} \sum  (\sigma_{\mathrm{IoU}}L_{\mathrm{cls}} + L_{\mathrm{reg}}) + \frac{1}{N_{\mathrm{neg}}} \sum L_{\mathrm{cls}}
 \label{eq:cls}
\end{equation}

\section{Experiments and discussions}\label{experiments}

\subsection{Dataset and evaluation metrics}

We evaluate our approach on the Team-OR dataset. As shown in Table~\ref{tab:split}, we split the dataset into a train set and test set, with 60\% and 40\% videos respectively. For evaluation metrics, we report the average precision (AP) at different temporal intersections over union (tIoU) thresholds, which is widely used in TAD research. 

\begin{table}[h]
\caption{Team-OR dataset split with the number of activities.}\label{tab:split}%
\centering
\resizebox{0.9\columnwidth}{!}{
\begin{tabular}{c|ccc}
\toprule
 & {Number of videos}  & Number of "Time-out" & Number of "StOP?"\\
\midrule
Train set   & 25  & 20  & 13  \\
Test set    & 18  & 13  & 9  \\
Total       & 43  & 33  & 22  \\
\bottomrule
\end{tabular}
}
\end{table}

\begin{table*}[h]
\caption{Comparison with the state-of-the-art TAD approaches. We report AP at different tIoU. }\label{tab:results}
\centering
\resizebox{0.8\textwidth}{!}{
\begin{tabular*}{\textwidth}{@{\extracolsep\fill}c|cccccc}
\toprule%
\multirow{2}{*}{Methods} & \multicolumn{6}{@{}c@{}}{Time-out} \\\cmidrule{2-7}%
 & 0.1 & 0.2 & 0.3 & 0.4 & 0.5 & Avg. \\
\midrule
ActionFormer \citep{zhang2022actionformer}  & 84.39 & 84.39 & 71.67 & 57.99 & 33.31 & 66.35 \\
TriDet~\citep{shi2023tridet}  & 98.90 & 98.90 & 98.90 & 92.86 & 65.63 & 91.04 \\
TemporalMaxer~\citep{tang2023temporalmaxer} & 93.05 & 93.05 & 93.05 & 93.05 & 93.05 & 93.05 \\
Ours  & \textbf{99.45} & \textbf{99.45} & \textbf{99.45} & \textbf{99.45} & \textbf{99.45} & \textbf{99.45} \\
\midrule
\multirow{2}{*}{Methods} & \multicolumn{6}{@{}c@{}}{StOP?} \\\cmidrule{2-7}%
 & 0.1 & 0.2 & 0.3 & 0.4 & 0.5 & Avg. \\
\midrule
ActionFormer~\citep{zhang2022actionformer}  & 12.63 & 12.63 & 12.63 & 5.79 & 1.22 & 8.98 \\
TriDet~\citep{shi2023tridet}  & 23.46 & 13.37 & 13.35 & 13.27 & 1.87 & 13.06 \\
TemporalMaxer~\citep{tang2023temporalmaxer} & 16.04 & 14.36 & 13.72 & 13.53 & 12.38 & 14.01 \\
Ours  & \textbf{30.85} & \textbf{20.78} & \textbf{20.66} & \textbf{20.58} & \textbf{20.52} & \textbf{22.68} \\
\bottomrule
\end{tabular*}
}
\end{table*}

\subsection{Implementation details}

For the video preprocessing, we resize the frames to 224*224 resolution, and we extract the clip features at a stride of 32 frames. Each clip contains 16 frames with an interval of 4 frames and 8 frames for VideoMAEv2~\citep{wang2023videomae} and STGCN++~\citep{duan2022pyskl} models respectively. We extract the features with a single NVIDIA A100 GPU.

We use the PyTorch framework to implement our approach and conduct experiments with a single NVIDIA A40 GPU. We train the model for 40 epochs with a batch size of 2. We use the AdamW optimizer with an initial learning rate of 1e-4. We set $L$ to 7 in Eq.~\ref{eq:max} and Eq.~\ref{eq:avg}.

\begin{table*}[h]
  \caption{Ablation study of the components of our approach, including using Max Pooling branch (Max.), Avg Pooling branch (Avg.) and skeleton feature. Baseline refers to the TriDet~\citep{shi2023tridet} model.}\label{tab:abl}
  \centering
\resizebox{0.8\textwidth}{!}{
    \begin{tabular*}{\textwidth}{@{\extracolsep\fill}c|ccc|cccccc}
        \toprule
        \multirow{2}{*}{Methods} & \multirow{2}{*}{Max.} & \multirow{2}{*}{Avg.} & \multirow{2}{*}{Skeleton} & \multicolumn{6}{@{}c@{}}{Time-out} \\\cmidrule{5-10} & & & & 0.1 & 0.2 & 0.3 & 0.4 & 0.5 & Avg. \\
        \midrule
        Baseline &  &  &  & 98.90 & 98.90 & 98.90 & 92.86 & 65.63 & 91.04 \\
        1 & \checkmark & & & \textbf{100.00} & \textbf{100.00} & \textbf{100.00} & \textbf{100.00} & 66.80 & 93.36 \\
        2 & \checkmark & \checkmark & & 98.97 & 98.97 & 98.97 & 98.97 & 79.06 & 94.99 \\
        3 & \checkmark & \checkmark & \checkmark & 99.45 & 99.45 & 99.45 & 99.45 & \textbf{99.45} & \textbf{99.45} \\
        \midrule
        \multirow{2}{*}{Methods} & \multirow{2}{*}{Max.} & \multirow{2}{*}{Avg.} & \multirow{2}{*}{Skeleton} & \multicolumn{6}{@{}c@{}}{StOP?} \\\cmidrule{5-10} & & & & 0.1 & 0.2 & 0.3 & 0.4 & 0.5 & Avg. \\
        \midrule
        Baseline &  &  &  & 23.46 & 13.37 & 13.35 & 13.27 & 1.87 & 13.06 \\
        1 & \checkmark & & & 25.72 & 23.69 & 23.65 & 7.43 & 7.26 & 17.55 \\
        2 & \checkmark & \checkmark & & 24.06 & \textbf{24.00} & \textbf{24.00} & 12.79 & 12.77 & 19.52 \\
        3 & \checkmark & \checkmark & \checkmark & \textbf{30.85} & 20.78 & 20.66 & \textbf{20.58} & \textbf{20.52} & \textbf{22.68} \\
        \bottomrule
    \end{tabular*}
    }
\end{table*}

\begin{table*}[h]
\caption{Ablation study of different features. Features with * are skeleton-based features. }\label{tab:feature}
\centering
\resizebox{0.8\textwidth}{!}{
\begin{tabular*}{\textwidth}{@{\extracolsep\fill}c|cccccc}
\toprule%
\multirow{2}{*}{Features} & \multicolumn{6}{@{}c@{}}{Time-out} \\\cmidrule{2-7}%
 & 0.1 & 0.2 & 0.3 & 0.4 & 0.5 & Avg. \\
\midrule
TC-CLIP~\citep{kim2025leveraging}  & 92.77 & 92.77 & 92.77 & 76.45 & 68.65 & 84.68 \\
VideoMAEv2~\citep{wang2023videomae}  & 98.97 & 98.97 & 98.97 & 98.97 & 79.06 & 94.99 \\
AGCN*~\citep{shi2019two} & 99.45 & 99.45 & 97.25 & 65.23 & 52.64 & 82.81 \\
PoseC3D*~\citep{duan2022revisiting}  & 96.26 & 92.98 & 92.98 & 82.15 & 67.84 & 86.44 \\
STGCN++*~\citep{duan2022pyskl} & 95.38 & 95.38 & 95.38 & 81.83 & 69.08 & 87.41 \\
\midrule
\multirow{2}{*}{Features} & \multicolumn{6}{@{}c@{}}{StOP?} \\\cmidrule{2-7}%
 & 0.1 & 0.2 & 0.3 & 0.4 & 0.5 & Avg. \\
\midrule
TC-CLIP~\citep{kim2025leveraging}  & 0.70 & 0.25 & 0.10 & 0.09 & 0.05 & 0.24 \\
VideoMAEv2~\citep{wang2023videomae}  & 24.06 & 24.00 & 24.00 & 12.79 & 12.77 & 19.52 \\
AGCN*~\citep{shi2019two} & 4.11 & 4.06 & 0.94 & 0.14 & 0.03 & 1.85 \\
PoseC3D*~\citep{duan2022revisiting}  & 1.13 & 0.92 & 0.88 & 0.46 & 0.45 & 0.77 \\
STGCN++*~\citep{duan2022pyskl} & 3.36 & 3.23 & 2.86 & 2.83 & 0.93 & 2.64 \\
\bottomrule
\end{tabular*}
}
\end{table*}

\subsection{Results}

Table~\ref{tab:results} compares our approach with the state-of-the-art TAD approaches, and it shows that the StOP? detection is much more difficult than the Time-out detection. In comparison, our approach performs the best on both activities, which establishes a solid baseline for this task. Also, it is worth mentioning that ActionFormer~\citep{zhang2022actionformer} has the most trainable parameters with the full self-attention, while TemporalMaxer~\citep{tang2023temporalmaxer} and ours approach only use Pooling in replacement but have better performance. This shows that the light-weight design of our model is important to prevent overfitting on our dataset, as we only have very few positive samples. 

Additionally, to test the applicability of our approach in a real-time setting, we conduct the real-time testing by computing the Frames Per Second (FPS) of the whole framework including feature extraction. The result shows that the overall FPS reaches 33.03, which is sufficient for real-time application.

\subsection{Ablation study}\label{sec:abl}

To evaluate the design of our framework including the Max Pooling and Avg Pooling branches and the skeleton-based features, we conduct an ablation study. According to Table~\ref{tab:abl}, the Pooling branches and the skeleton features are all necessary for better performance. It is also noticeable that the Avg Pooling branch and the skeleton features improve the boundary regression, leading to a higher AP@0.5. 

Additionally, we conduct an ablation study of different video features. As shown in Table~\ref{tab:feature}, VideoMAEv2 performs significantly better than any other model, proving the importance of video representations with high granularity, as VideoMAEv2 utilizes a masking strategy with extremely high ratio (90\% to 95\%) for self-supervised learning. Although skeleton models perform decently on the Time-out detection, they perform poorly on the StOP? detection. In general, an in-depth study is needed to develop a more robust and fine-grained action representation in the OR.

\subsection{Limitations and future work}

In this work, although we have studied the automatic detection of standard team communication protocols in the OR for the first time, there are some limitations: the evaluation is only conducted on one clinical site, the number of protocol samples is limited, the detection of the StOP?-protocol remains challenging, and the more complex spontaneous team interactions remain unexplored. 


In the future, we envision that our approach could be deployed across various clinical settings. The standardized instructions of both the Time-out and StOP?-protocols ensure consistent implementation across hospitals, while the lightweight design of our model facilitates training with limited data availability. Additionally, we aim to investigate more fine-grained, spontaneous individual interactions to explore potential connections between these atomic-level interactions and team communication protocols, ultimately enhancing the robustness of interaction analysis in the OR.

\section{Conclusion}\label{conclusion}

In this paper, we take the first step towards understanding team interactions in the OR, by studying the clinically validated Time-out and StOP? activities during surgery. In particular, we introduce the first OR dataset of real surgeries, which is of great value for team communication analysis. We also propose a challenging task of localizing the Time-out and StOP?-protocol activities given untrimmed OR videos and design a novel group activity detection algorithm as a solid baseline. 


%







\section{Acknowledgements}

This work was supported by the Sherbard Foundation, by French state funds managed by BPI France (project 5G-OR) and by the ANR under reference ANR-10-IAHU-02 (IHU Strasbourg). This work was also granted access to the servers/HPC resources managed by CAMMA, IHU Strasbourg, Unistra Mesocentre, and GENCI-IDRIS [Grant 2021-AD011011638R3]. We thank Andreas Weibel and Lionel Bergerot for setting up the recording system in the operating room and for providing technical support throughout the study. We thank Laurin Terhorst for his contribution to the data collection. 
\\

\textbf{Ethical approval:} The project is conducted under the following protocols: the Swiss legal requirements, the current version of the World Medical Association Declaration of Helsinki, the principles and procedures for integrity in scientific research involving human beings, the {\it StOP? II} clinical trial and the study ``Assessment of team coordination in the operating room based on motion analysis: A proof of concepts study''.

\textbf{Competing interests:} The authors declare no conflict of interest. 

\textbf{Informed consent:} Data was collected with informed consent from the human participants involved.

\bibliographystyle{model2-names.bst}
\bibliography{sn-bibliography}

\end{document}